\documentclass[10pt,twocolumn,letterpaper]{article}

\usepackage{cvpr}
\usepackage{times}
\usepackage{epsfig}
\usepackage{graphicx}
\usepackage{amsmath}
\usepackage{amssymb}

\usepackage[breaklinks=true,bookmarks=false]{hyperref}

\cvprfinalcopy 

\def\cvprPaperID{****} 

\begin{document}
\title{EfficientNet-eLite: Extremely Lightweight and Efficient CNN Models for Edge Devices by Network Candidate Search}

\author{Ching-Chen Wang, \ \ Ching-Te Chiu, \ \ Jheng-Yi Chang  \\
Department of Computer Science, National Tsing Hua University.\\
{\tt\small jason01200120@gapp.nthu.edu.tw, chiusms@cs.nthu.edu.tw, s108064529@m108.nthu.edu.tw}
}
\maketitle

\begin{abstract}
Embedding Convolutional Neural Network (CNN) into edge devices for inference is a very challenging task because such lightweight hardware is not born to handle this heavyweight software, which is the common overhead from the modern state-of-the-art CNN models. In this paper, targeting at reducing the overhead with trading the accuracy as less as possible, we propose a novel of Network Candidate Search (NCS), an alternative way to study the trade-off between the resource usage and the performance through grouping concepts and elimination tournament. Besides, NCS can also be generalized across any neural network. In our experiment, we collect candidate CNN models from EfficientNet-B0 to be scaled down in varied way through width, depth, input resolution and compound scaling down, applying NCS to research the scaling-down trade-off. Meanwhile, a family of extremely lightweight EfficientNet is obtained, called EfficientNet-eLite.

For further embracing the CNN edge application with Application-Specific Integrated Circuit (ASIC), we adjust the architectures of EfficientNet-eLite to build the more hardware-friendly version, EfficientNet-HF. Evaluation on ImageNet dataset, both proposed EfficientNet-eLite and EfficientNet-HF present better parameter usage and accuracy than the previous start-of-the-art CNNs. Particularly, the smallest member of EfficientNet-eLite is more lightweight than the best and smallest existing MnasNet with 1.46x less parameters and 0.56\% higher accuracy. Code is available at \url{https://github.com/Ching-Chen-Wang/EfficientNet-eLite}
\end{abstract}


\begin{figure}[t]
\begin{center}

   \includegraphics[width=\linewidth]{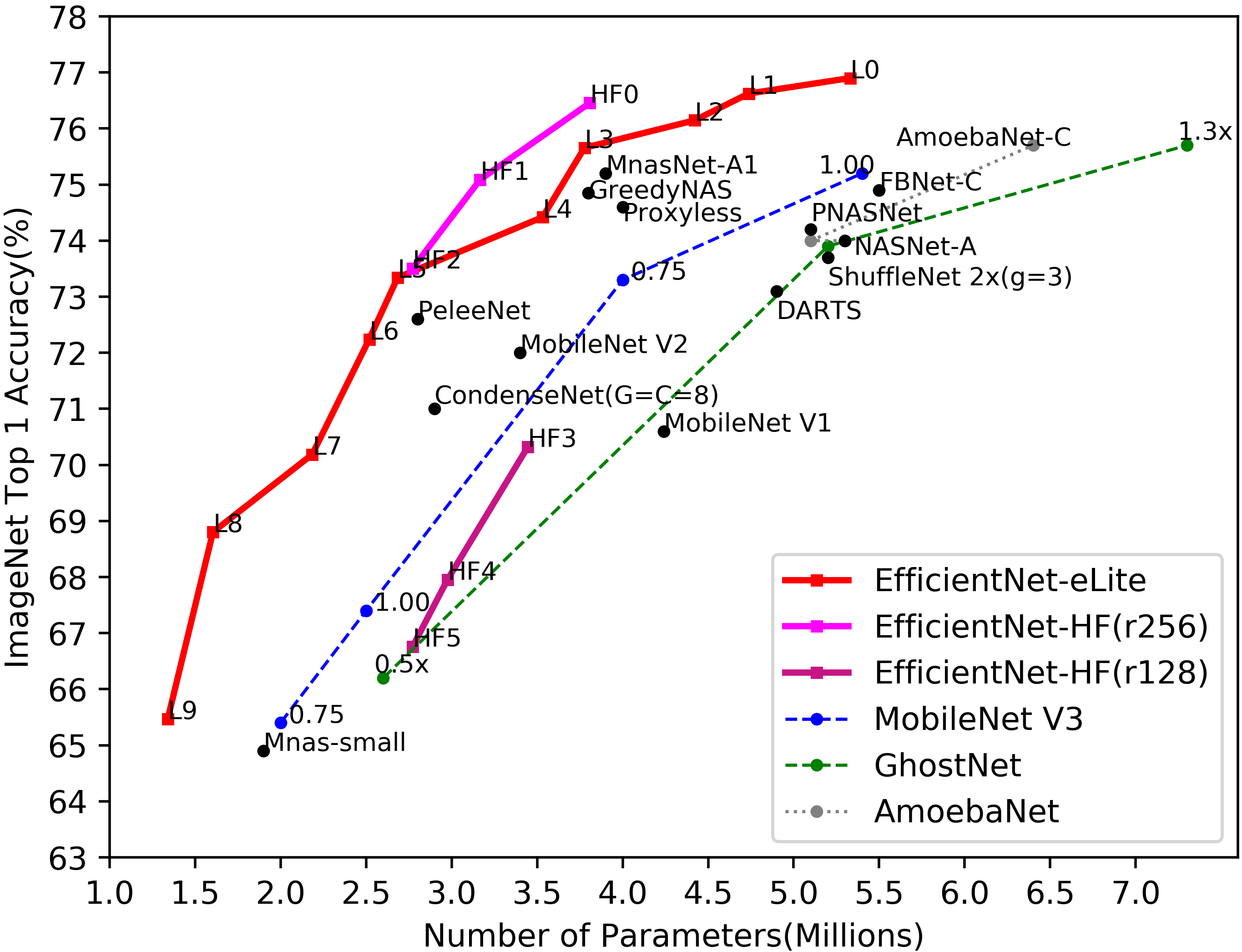}
\end{center}
   \caption{The performance of model size and Top-1 accuracy on ImageNet \cite{ImageNet}. The proposed CNN family, EfficientNet-eLite, is more lightweight than the other state-of-the-art models with higher accuracy on ImageNet dataset. Particularly, the smallest member of EfficientNet-eLite is more lightweight than the best and smallest existing MnasNet with 1.46x less parameters and 0.56\% higher accuracy. As for the two versions of hardware-friendly models, r128 and r256 denote the input resolution 128x128 and 256x256 respectively. More details about performance are provided in Table \ref{table:state}.}
\label{fig:Parameter}
\end{figure}

\section{Introduction}

In recent decade, Convolutional Neural Network (CNN) presents the remarkable achievement on vision task such as action recognition \cite{Action}, object detection \cite{Obj1}\cite{Obj2}, image segmentation \cite{image_segmentation} and so on \cite{optical_flow}\cite{image_captioning1}\cite{image_captioning2}. However, the CNN-based application is still uncommon nowadays, which we attribute the reasons to the hardware limitation and the software complexity. The former one is that the application is strictly limited by the hardware condition. Enough RAM to save the heavy parameters and powerful computational ability to perform tons of operations are both necessary requirement of the CNN-based application. Besides, portability and some physical limitations are also preventing its development. On the other hand, the modern outperformed CNN models are always featured with the intensive computation and parameters. Although the mobile-size CNN models are proposed recently~\cite{MobileNets}~\cite{MobileNets_v2}~\cite{MobileNets_v3}, the CNN models are not lightweight enough for some edge devices such as low-power IoT (Internet of things) devices or wearable device such as smart glasses, watches and so on.

The CNN accelerators appear to bridge the gap between the CNN applications and the edge devices~\cite{eCNN}~\cite{RGBD_eCNN}. Nevertheless, apart form designing CNN on general-purpose hardware, the design space of CNN models on accelerator is scarified by some degrees of freedom and strictly restricted by the hardware specification. To be more specific, all types of operation in CNN should be fully compatible according to the instruction set architecture (ISA). In addition to the compatibility, the performance of ASIC is another important design principle such as chip area, energy efficiency, utilization and so on. In ~\cite{eCNN}, the state-of-the-art methods are highlight for the the codesign concepts of the software and the hardware, to build not only accurate but also hardware-friendly CNN structure.

\begin{figure}[t]
\begin{center}

   \includegraphics[width=\linewidth]{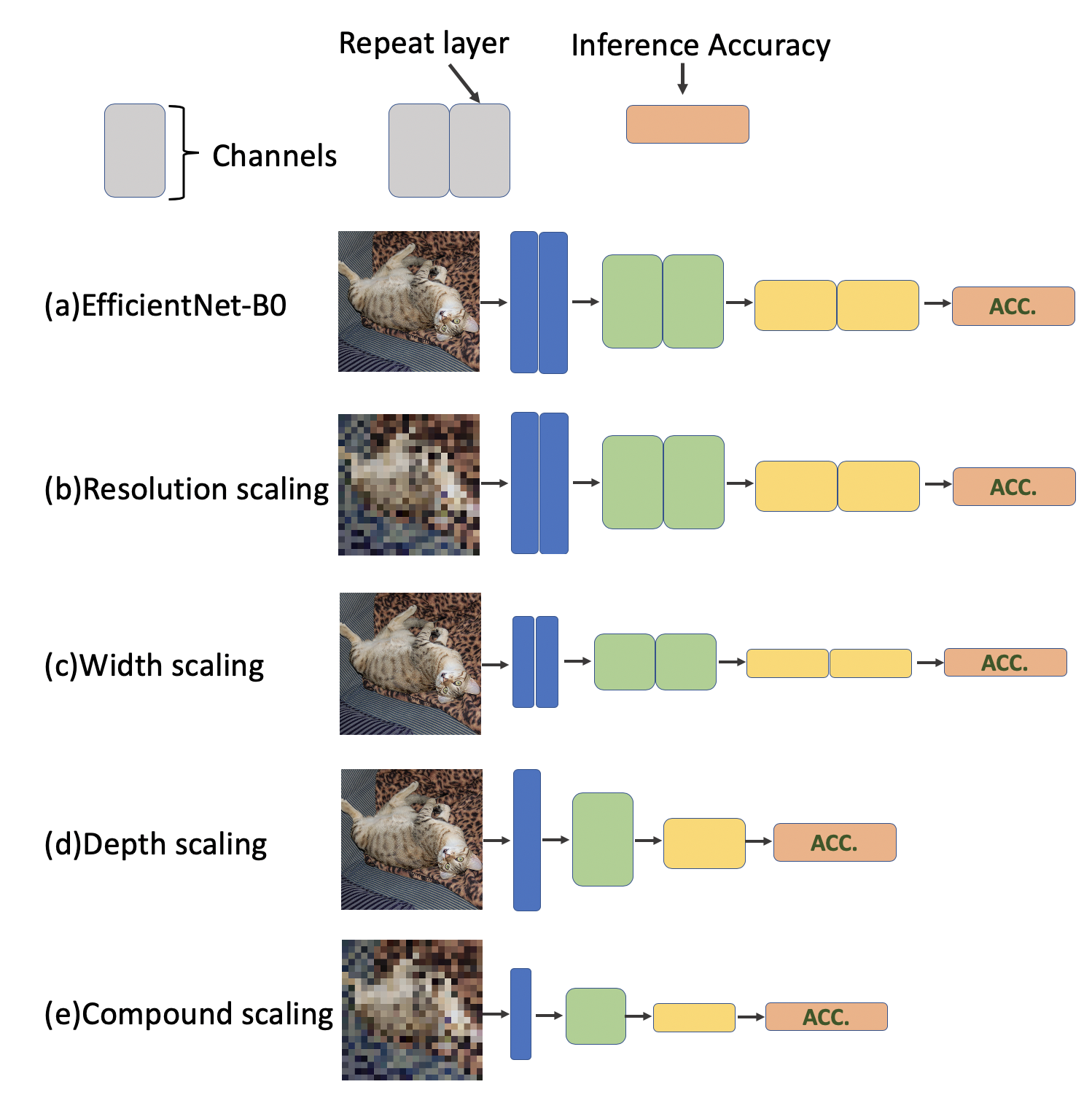}
\end{center}
   \caption{We make EfficientNet more lightweight by scaling down EfficientNet-B0 (a) through Resolution (b), Width (c), Depth (d) and Compound (e).}
\label{fig:long}
\label{fig:scaling}
\end{figure}

In this study, the modern outperformed CNN models, EfficientNets~\cite{EfficientNet}, are selected as our backbone structure. We target at building more lightweight version and adjusting toward a more hardware-friendly structure. First of all, we apply EfficientNet-B0 (Baseline model) to be {\bf scaled down} among channels, depths, input resolution as a technique of model compression, along with the compound scaling method and the constant ratio proposed by ~\cite{EfficientNet}, shown in the Figure \ref{fig:scaling}. The thinner EfficientNets are collected into candidate pool and used for studying the trade-off by proposed Network Candidate Search (NCS). Although compound scaling up with fixed scaling coefficients is systematically analyzed in ~\cite{EfficientNet}, the {\bf scaling down} principle is not well understand. We believe that those three dimensions may possibly not keep the scaling relationship of constant ratio. Our study indicates that input information plays more significant role than channels and depths when a model is scaled down. 

\begin{figure}[t]
\begin{center}

   \includegraphics[width=\linewidth]{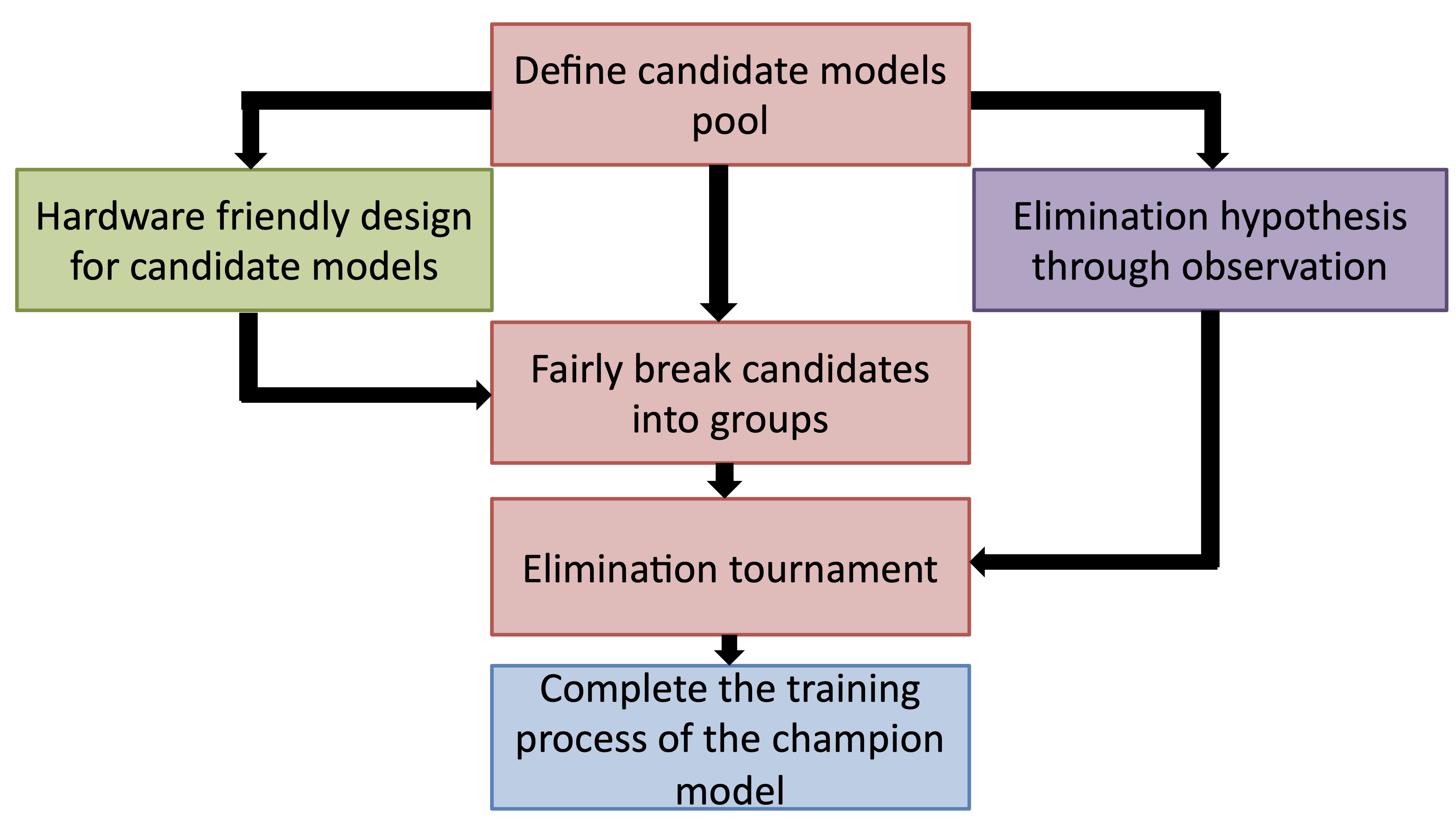}
\end{center}
   \caption{The overview of proposed Network Candidate Search.}
\label{fig:long}
\label{fig:NCS}
\end{figure}

The Figure \ref{fig:NCS} provides the overview of NCS, which is composed of two major concepts, grouping and elimination tournament respectively. The former is trying to investigate the question : what is the CNN's shape (i.e. the ratio or relationship of width, depth and input resolution) that can achieve better accuracy. The basic idea is that candidates (CNN models) with similar parameter usage and Flops are fairly divided into group for comparing. By doing so, the outperformed models inside the group have the relatively better shape and the worthy scaling-down trade-off. The letter is to mitigate the training cost under an affordable GPU hours. Only the potential models in each group survive and the others gradually eliminate as a method to stop the training for releasing the burden of GPU. As for determining the potential models, we adopt using average accuracy as criteria, coming form observing the relationship between learning performance and final accuracy. Finally the champion of the elimination tournament in each group is obtained, called EfficientNet-eLite (Extremely lightweight EfficientNet), which presents better parameter usage and accuracy than the previous start-of-the-art CNNs. Particularly, our EfficientNet-eLite 9 outperforms MnasNet with 1.46x less parameters and 0.56\% higher accuracy on ImageNet.

Secondly, to go further alleviating the difficulty of the CNN inference on the edge, we provide the  hardware-friendly CNN models as candidates for NCS by considering design concepts of Application-Specific Integrated Circuit (ASIC). Finally, we obtain a family of relatively outperformed models, called EfficientNet-HF (Hardware-friendly EfficientNet), realizing CNN models could be not only accurate but also hardware-friendly for ASIC.

The rest of this paper is organized as follow. Section 2 presents the related works, including scaling methods that we apply for model compression and our baseline model EfficientNet~\cite{EfficientNet} as well as some hardware-friendly designs for CNN. Section 3 discusses the proposed Network Candidate Search in detail. Proposed family of Hardware-friendly EfficientNet is introduced in Section 4. Experimental results and conclusion are depicted in Section 5 and Section 6 respectively.


\section{Related Work}
\label{section:Model_scaling}
\subsection{Model scaling}
\noindent {\bf  Model scaling} has been a very popular method for expanding the scale of CNN models to pursue the better accuracy. The early CNN, LeNet-5 for recognizing handwriting digits, has only 7 layers and thousands of trainable parameters. With the progress of the Graphics Processing Unit (GPU), CNN has evolved to solve the more complicated problems by going bigger and bigger. AlexNet~\cite{AlexNet}, the deeper network with 8 layers, is the widely known breakthrough in 2012 ImageNet Large Scale Visual Recognition Challenge (ILSVRC) competition with about sixty millions  parameters. VGG16~\cite{Vgg16}, 16 layers and regular architecture, propose a consecutive stack of two 3x3 convolution layers to replace a single 5x5 convolution layer, which makes its structure more deeper. ResNet~\cite{ResNet} family wins the 2015 ImageNet challenge. The vanishing gradient problem has been dealt with by ResNet to achieve 152 layers architecture.

We categorize the common ways to adjust the model size into 4 segments, which are introduced respectively as following.

\noindent {\bf Depth scaling :}
It is widely believed that deeper network should achieve better performance ~\cite{ResNet}~\cite{Deep_Networks}~\cite{Going_deeper}~\cite{Rethinking_the_inception}. Once the vanishing gradient problem is dealt the network depth can be deeper and gain the accuracy by tuning the repeat time of the basic building blocks. 
For example, by modifying the repetition of residual blocks, ResNet can be scaled up to 152 layers or scaled down to only 18 layers.

\noindent {\bf Width scaling : }
Several state of the arts commonly apply width scaling~\cite{MobileNets}~\cite{MobileNets_v2}. The intuition is that wider network has more filters to memorize the input patterns. Besides, wider network has the tendency of making the training easier~\cite{Wide_residual_networks}.

\noindent {\bf Resolution scaling : }
Higher resolution can provide more detail information for CNN models to tell apart from the difference between the very similar input. Therefore, the accuracy can be improved. Beginning from input size 224 x 224, Inception-V4 applies 299 x 299 as input size. A larger input size, 480 x 480, is used in ~\cite{Gpipe}.

\noindent {\bf Compound scaling : }
With the CNN scaled up to reach the hardware limit, researchers are devoting to finding the more efficient way of scaling. To be more specific, how the additional hardware resource should be effectively assigned into the scaling dimension becomes an important question.
That is to say, the scaling trade-off becomes an active research targeting how to gain accuracy with less resource cost. EfficientNet\cite{EfficientNet} authors conduct series of experiment to find out the observation that scaling through single dimension will quick saturate the accuracy gain and that compound scaling through three dimensions of depth, width and input resolution will achieve better performance.


\subsection{EfficientNet}
\label{section:EfficientNet}

\begin{table}

	\vskip -0.1in     
\caption{Family of EfficientNet by scaling up.}     
  \vskip 0.05in
  \centering   

  \resizebox{1.0\columnwidth}{!}{ 
	    \begin{tabular}{|c|c|c|c|c|c|c|}
        \hline
        {EfficientNet} & {B0} & {B1}  & {B2}& {B3} & {B4}  & {...}\\
        \hline
        $\phi $(Available resource)  & 0&-&1&2&3&...\\
        \hline
        Depth & $D$ & 		$1.1 \cdot D$	 &  $1.2 \cdot D$      &       $1.2^2 \cdot D$      &   $1.2^3 \cdot D$   & $d \cdot D$\\
        \hline
        Width & $W$ & 		$W$	 &  $1.1 \cdot W$      &       $1.1^2 \cdot W$      &   $1.1^3 \cdot W$   & $w \cdot W$\\
        \hline
        Resolution  & $R$ & 		$1.07 \cdot R$	 &  $1.15 \cdot R$      &       $1.15^2 \cdot R$      &   $1.15^3 \cdot R$   & $r \cdot R$\\
        \hline
        Parameters  & $5.3M$ & 		$7.8M$	 &  $9.2M$      &       $12M$      &   $19M$   & $...$\\
         \hline
        Flops  & $0.39B$ & 		$0.70B$	 &  $1.0B$      &       $1.8B$      &   $4.2B$   & $...$\\
        \hline
    \end{tabular}
                                     
   }                                                                                 
	
  \label{tab:efficientnetb0}      
\end{table}

\begin{figure}[t]
\begin{center}

   \includegraphics[width=0.8\linewidth]{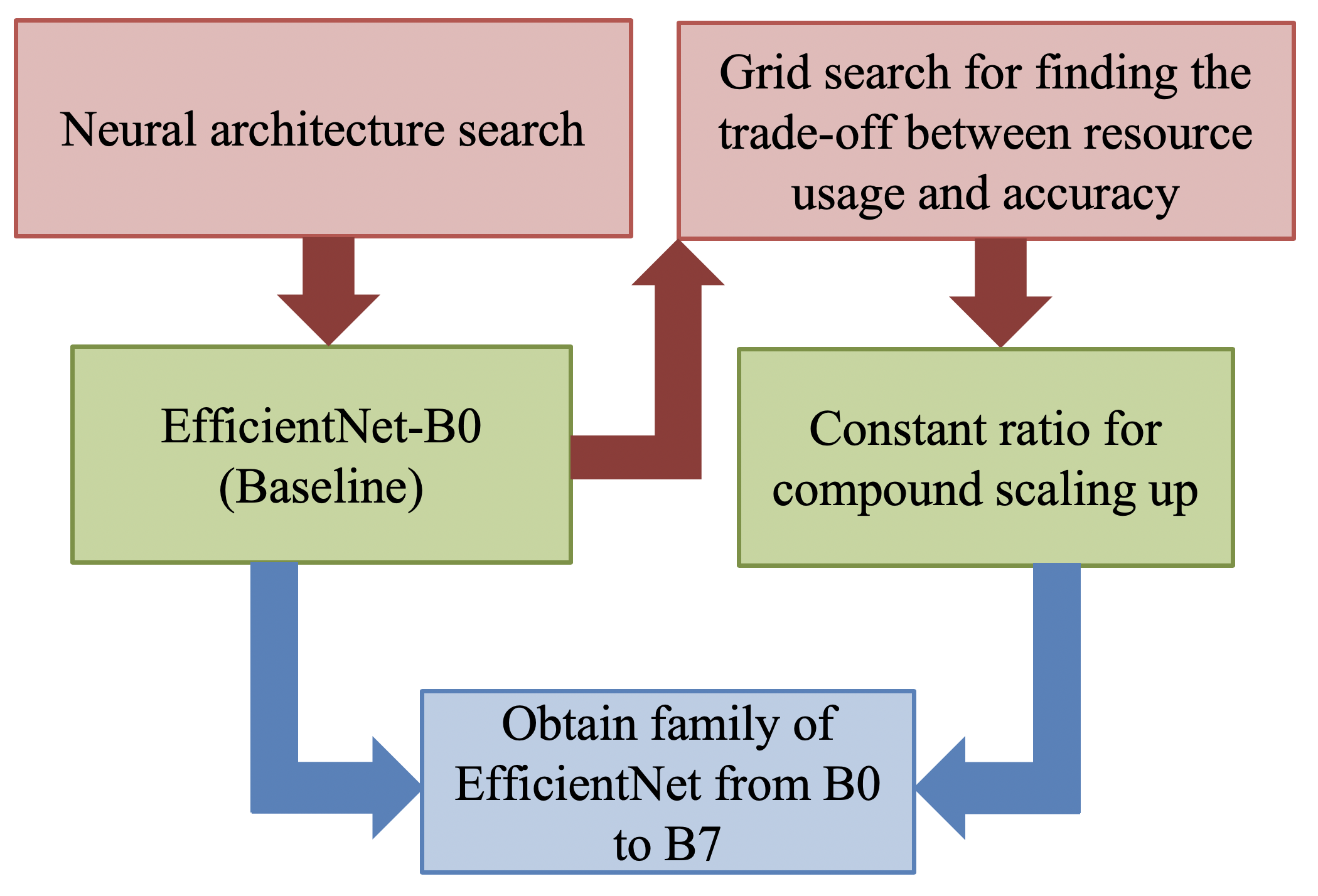}
\end{center}
   \caption{The overview of EfficientNet.}
\label{fig:long}
\label{fig:EfficientNet_flow}
\end{figure}

 The Figure \ref{fig:EfficientNet_flow} illustrates the overview of building up family of EfficientNet and we divide the state-of-the-art EfficientNet into two parts to discuss.

\begin{table*}
	\vskip -0.1in     
	\caption{The structure of the original and the scaled-down EfficientNet.}      
  \vskip 0.05in
  \centering   

  \resizebox{1.0\textwidth}{!}{ 
\begin{tabular}{|c|c|c|c|c|c|c|c|c|c|c|c|c|} 
\hline

Stage &Operator&Resolution &Resolution&Channels&Channels&Repeat &Repeat & \multicolumn{5}{c|}{Scaled down results of repeat time of operator}  \\ 
	 &		&(Baseline)  &(Scaled down)  &(Baseline) &(Scaled down)&(Baseline)&(Scaled down)&\multicolumn{1}{c}{}&\multicolumn{1}{c}{}&\multicolumn{1}{c}{}&\multicolumn{1}{c}{}&\multicolumn{1}{c|}{} \\ 
$s$ & & $H_s \times W_s$ & $H_s r_k \times W_s r_k$ & $C_s$ & $C_s \times w_i$ & $ R_s$ & $  R_s \times d_j$&\multicolumn{1}{c}{$d_1 = 1.0$}&\multicolumn{1}{c}{$d_x = 0.9 \ or \ 0.8$} &\multicolumn{1}{c}{$d_2 = 0.7$} &\multicolumn{1}{c}{$d_3 = 0.6$} &\multicolumn{1}{c|}{$d_4 = 0.5$} \\ \hline

1&Conv3x3&$224 \times 224$&$224r_k \times 224r_k$&$32$&$32 \times w_i$&$1$&$1 \times d_j$&1&1&1&1&1\\ \hline

2&MBC1, 3x3&$112 \times 112$&$112r_k \times 112r_k$&16&$16 \times w_i$&1&$1 \times d_j$&1&1&1&1&1 \\ \hline

3&MBC6, 3x3&$112 \times 112$&$112r_k \times 112r_k$&24&$24 \times w_i$&2&$2 \times d_j$&2&2&2&2&1 \\ \hline

4&MBC6, 5x5&$56 \times 56$&$56r_k \times 56r_k$&40&$40 \times w_i$&2&$2 \times d_j$&2&2&2&2&1\\ \hline

5&MBC6, 3x3&$28 \times 28$&$28r_k \times 28r_k$&80&$80 \times w_i$&3&$3 \times d_j$&3&3&3&2&2\\ \hline

6&MBC6, 5x5&$14 \times 14$&$14r_k \times 14r_k$&112&$112 \times w_i$&3&$3 \times d_j$&3&3&3&2&2\\ \hline

7&MBC6, 5x5&$14 \times 14$&$14r_k \times 14r_k$&192&$192 \times w_i$&4&$4 \times d_j$&4&4&3&2&2\\ \hline

8&MBC6, 3x3&$7 \times 7$&$7r_k \times 7r_k$&320&$320 \times w_i$&1&$1 \times d_j$&1&1&1&1&1\\ \hline

9&AVG, FC&$7 \times 7$&$7r_k \times 7r_k$&1280&$1280 \times w_i$&1&$1 \times d_j$&1&1&1&1&1\\ \hline
\end{tabular}                                                            
   }

  \label{table:structure}  
\end{table*} 
\noindent {\bf Neural Architecture Search : } 
Neural Architecture Search (NAS) is a technique based on reinforcement learning to automatically build up CNN models. The conventional CNN structure is made in hand-crafted manner, which has lots of architectural possibilities and significantly relies on human's expertise. NAS mitigates the efforts of trial and error on constructing CNN networks and modern CNN models designed by automated approaches tends to outperform the manually designed one \cite{MnasNet}.

\noindent {\bf Grid Search : } The goal of grid search is to find out a strategy to effectively assign available hardware resources into depth $d$, width $w$ and input resolution $r$ for expanding CNN models. In this case, the authors assume twice more resources available, denoting $\phi =1$. Candidate scaling coefficients of depth $\alpha$, width $\beta$, and resolution $\gamma$ to determine how to allocate those resources. There are lots of combinations of $\alpha$, $\beta$ and $\gamma$. After the grid search, the authors find the best value for EfficientNet-B0 are $\alpha=1.2$, $\beta=1.1$ and $\gamma=1.15$. The EfficientNet family, shown in Table \ref{tab:efficientnetb0}, can be obtained by $d$, $w$ and $r$ according to the amount of available resources $\phi$.

\subsection{Hardware-friendly designs for CNN models}
\label{section:Hardware-oriented}

There are several hardware-friendly designs for CNN models. For compressing toward small size of model by using low precision data format, quantization and dynamic fixed-point representation are manipulated in ~\cite{eCNN}~\cite{dynamic_fix_point1}~\cite{dynamic_fix_point2}. Even binary precision is adopted in \cite{binary_network}. As for abolishing the redundant parameter, pruning and value decomposition are commonly used methodology. On the other hand, reducing the computational overhead is another type of hardware friendly principle. Depth-wise separable convolution is introduced in MobileNet~\cite{MobileNets}, known as impressively decrease
of operation. Hardware friendly activation functions~\cite{HF_activation}\cite{Hard-sigmoid} alleviate the challenge of fixed point arithmetic.  

\noindent {\bf Hardware-friendly CNN structure : }  Under general purpose hardware, the modern CNNs are optimized for parameter usage and Flops. As for the dedicated CNN accelerator, a regular and modularized architecture is considered as hardware-friendly design. Taking the state-of-the-art embedded CNN in \cite{RGBD_eCNN} as example, we realize that the structure is hardware-friendly for ASIC design regarded from channels and size of feature map in each layer. From the perspective of channels, the whole CNN is built upon based on 3x3 convolution with 32 channels input to 32 channels output. Therefore, the accelerator can have high utilization performance when process element (PE) is designed for performing the parallel 32 channels operations. For size of feature map to be power of two, the tiling technique and data partitioning are more easily to apply when the SRAM is not enough, thus to access the DRAM. Besides, the operation of division (average pooling) can be done by shifting rather than the hardware division unit.




\section{Network Candidate Search}
\label{section:NCS}
The core concept of NCS is searching for outperformed models over the candidate models which consume the similar hardware cost. In this section, we start from defining candidates with
EfficientNet-B0~\cite{EfficientNet} to be scaled down in varied way. Secondly, the similar candidates are grouped together for comparing. Thirdly, we introduce the criteria for elimination. Lastly, the algorithm of NCS is summarized.
 
\subsection{Candidate of CNN models}
The baseline CNN model is broken down into nine stages with relative operators~\cite{MnasNet}, listed in the Table \ref{table:structure}. Channels $C_s$, Resolution  $H_s \times W_s $ and Depth (sum of Repeat $R_s$) are denoted as the baseline specification with relative stage $s$. We preserve all types of operation inside the Operator, focusing on scaling down channels $w_i$, depth $d_j$ and input resolution $r_k$ such that $0 < w_i, d_j,r_k \leq1$, making the scaled down model more lightweight with Channels $C_s \times w_i$, Resolution  $H_s r_k \times W_s r_k$ and Repeat $R_s \times d_j$.

We define a candidate pool $CP$ as a set, and each element in the pool symbolizes a CNN model, which is determined by various combination of scaling coefficients $w_i, d_j, r_k$. Values $i,j,k$ denote as index representing the magnitude of scaling.

\begin{equation} \label{eq:CP} 
\text{$CP$: }  \{ Model(w_i, d_j, r_k) | \forall i,j,k \in  {\mathbb{N}}  \ and \  0<w_i, d_j,r_k \leq 1 \} \\
\end{equation}

\noindent {\bf Define Scaling coefficient from depth : }We find that depth coefficient is easier to determine than width and input resolution because repeat time has fewer possibilities. For example, original input resolution is $224 \times 224$. The resolution to be scaled down could be $223 \times 223$, $222 \times 222$ ... and so on. Channel coefficients have lots of choices for the same reason. However, depth coefficients have just few cases, which we can start defining from. The right side of Table \ref{table:structure} illustrates the scaled down results of depth. In Table \ref{table:structure}, $d_x$ means the "don't care". Because after the ceiling function of repeat time, applying the coefficients 0.9 and 0.8 will result the same depth as $d_1 = 1.0$. Thus, we define $d_2=0.7$ and  $d_3, d_4, ..., d_j$ according to Equation \ref{eq:d_j}.

\begin{equation} \label{eq:d_j} 
\begin{aligned}
\text{$d_1$} = &1.0  \ if j =1, \\
\text{$d_j$} = &  d_{j-1}  - 0.1x, \ \exists x \in  {\mathbb{N}}, \\
s.t. \  \displaystyle\sum_{s=1}^{9} ceiling(R_s \cdot d_{j-1})& > \  \displaystyle\sum_{s=1}^{9} ceiling(R_s \cdot d_{j}) \ if j > 1\\
\end{aligned}
\end{equation}

\begin{table}
	\vskip -0.1in     
\caption{Coefficients for scaling down EfficientNet.}    
  \vskip 0.05in
  \centering   

  \resizebox{0.7\columnwidth}{!}{ 
\begin{tabular}{|c|c|c|c|c|c|}
\hline
$d_j$            & $d_1$ & $d_2$  & $d_3$ & $d_4$ & $d_j$ \\ \hline 
Coefficient      & 1.0   & 0.7    & 0.6   & 0.5   & ...   \\ \hline
Total operators  & $t_1$ & $t_2$  & $t_3$ & $t_4$ & $t_u$   \\ \hline
                 & 18    & 17     & 15    & 12    & ...   \\ \hline \hline
$w_i$            & $w_1$ & $w_2$  & $w_3$ & $w_4$ & $w_i$ \\ \hline
Coefficient      & 1.0   & 0.8666 & 0.701 & 0.514 & ...   \\ \hline \hline
$r_k$            & $r_1$ & $r_2$  & $r_3$ & $r_4$ & $r_k$ \\ \hline
Coefficient      & 1.0   & 0.905  & 0.766 & 0.587 & ...   \\ \hline
Input resolution & 224   & 203    & 172   & 132   & ...   \\ \hline
\end{tabular}
   }                                                                                 
	 
\label{table:result_of_scaling}  
\end{table}

\noindent {\bf Define width and resolution coefficients : }
EfficientNet \cite{EfficientNet} scales up model from B0 to B7 by a set of constant ratio $(w = 1.1, d = 1.2_j, r = 1.15)$, which is obtained by the grid search under the predefined resource budget. We already have the depth coefficients $d_j$. The idea is that we can use this set of constant ratio to scale down, calculating corresponding $w_i, r_k$ with the depth coefficients with Equation \ref{eq:scaling_detail}.
Note that we use total amount of operators $t_u$ instead of $d_j$ because $t_u$ is more representative for the depth coefficient in EfficientNet \cite{EfficientNet} and that we only consider $i, u, k$ less or equal than 4 due to the GPU resource limitation. 

Instead of directly using the compound scaling coefficients (i.e. Model$(w_2, d_2,r_2)$, Model$(w_3, d_3,r_3)$ or Model$(w_4, d_4,r_4))$, the flexibility is considered by collecting all the combination of $w_i, d_j, r_k$ into candidate pool. By this way, we can investigate the shape of CNN (i.e. the ratio or relationship of width, depth and input resolution) through the different scaled-down strategies, studying the scaled-down trade-off along with the proposed grouping method, introduced in the following section. 

\begin{equation} \label{eq:scaling_detail} 
\begin{aligned}
&\text{$\bigtriangleup w$} = \dfrac{w_{i+1}}{w_i}, \ \text{$\bigtriangleup d$} = \dfrac{t_{u+1}}{t_u}, \ \text{$\bigtriangleup r$} = \dfrac{r_{k+1}}{r_k} \\
& \bigtriangleup w :  \bigtriangleup d :  \bigtriangleup r  = 1.1 : 1.2 : 1.15 
\end{aligned}
\end{equation}

\subsection{Grouping method} 
The core idea of grouping method is to research the different shapes of CNN under the similar kinds of resource consumption, which specifies as parameter usage and Flops of a CNN model. Namely, CNN models with similar parameter usage and Flops are gathered in the same group so that the outperformed one in the group can be considered as the relatively good shape of the CNN. Besides, the combination of scaling coefficients from width, depth and input resolution are the comparatively better strategy for scaling down.

For fairly comparing the different shapes of CNN in candidate pool, we keep the factors, which may affect the performance of CNN, the same as much as possible, including the training environment such as batch size, learning rate, data augmentation policy, optimization algorithm and so on. As for hardware resource usage, it is difficult to have candidates with the exactly same parameter usage and Flops.  Additionally, those two measurements are not in the same scope, making it more challenge to fairly group the candidates. The statistic values are provided as follow. For each model in candidate pool, the mean of parameter usage is  $\bar{X}_{Para} = 3.1$ million but the mean of Flops is  $\bar{X}_{Flops} = 153.4$ million. As for the dispersion of the candidate pool, standard deviation of parameter usage is $ {\sigma_{Para} = 1.2}$ million and the flops is $ {\sigma_{Para} = 90.8}$ million.

As a result, we propose a grouping method based on statistic distribution of both parameter usage and Flops. The benefit is that the statistic distribution is not affected by the scale. The Z-score value is calculated from each model in candidate pool. The Z-score is the offset of how many standard deviations from the mean the data is, denoting in Equation \ref{eq:Z-score}. Hence, with this offset, the larger Z-score value represents the relatively heavy resource cost and vice versa. The parameter usage and Flops are standardized to the same scale so that we can adopt $Z_{sum}$ as standard to classify the candidate models into groups. 

\begin{equation} \label{eq:Z-score} 
\begin{aligned}
\text{$Z_{Para} \ $= } & \ \dfrac{X_{Para}  - \bar{X}_{Para} }{\sigma_{Para} },\\
\text{$Z_{Flops}\ $=} &  \ \dfrac{X_{Flops}  - \bar{X}_{Flops} }{\sigma_{Flops} },\\
\text{$Z_{sum}\ $=} & \ Z_{Para} + Z_{Flops}\\
\end{aligned}
\end{equation}

\subsection{Criteria for elimination}
\label{section:Criteria for elimination}
Expensive searching cost is a common issue of the searching-based approaches. It is impractical to finish the training process of all candidate models. Therefore, several searching-based state-of-the-art methods apply the elimination strategy to alleviate the searching cost. The state-of-the-art MnasNet\cite{MnasNet} adopts the accuracy of fifth epoch to eliminate candidates during searching over 8,000 models. Other state-of-the-art methods, PNAS\cite{PNAS} applies the accuracy of $20^{th}$ epoch and AmoebaNet\cite{AmoebaNet} uses the accuracy of $25^{th}$ epoch to speculate whether the candidate is the outperformed CNN model. 

In this section, our goal is to find out the criteria to seek up potential models with the acceptable searching cost. Instinctively, the outperformed model should show its talent during the early training phase. Therefore, with few epochs of training, the outperformed model can be distinguished with higher accuracy in the early training phase. However, in our study, we realize that the candidates in each group have the closing accuracy during the early training phase, which may lead to mistakenly eliminate the promising model due to the unexpected result of specific epoch. As a result, we propose using averaging accuracy, which is counted the accuracy from the start to the current epoch, as criteria for elimination.

The thought is coming from the observation of the learning performance. First of all, we begin with k models, which have similar hardware resource usage. Due to GPU limitation, we set k=4 and the parameter usage and Flops are listed in the table \ref{table:Cmw}. We are targeting at finding the clues to know the ranking early (i.e. few training cost) rather than waiting until the final epoch. For example, in the table \ref{table:Cmw}, we hope to stop training C1 and C3 as early as we can, because C2 and C4 are the relatively outperformed models inside this group.

\noindent {\bf Observation : }The accuracy curves are crossing. (Before $50^th$ epoch, C2 and C3 have closing accuracy. C2 and C4 have crossing accuracy curve.)

\noindent {\bf Hypothesis : } Average accuracy is more representable for the final performance

\begin{figure}[t]
\begin{center}

   \includegraphics[width=\linewidth]{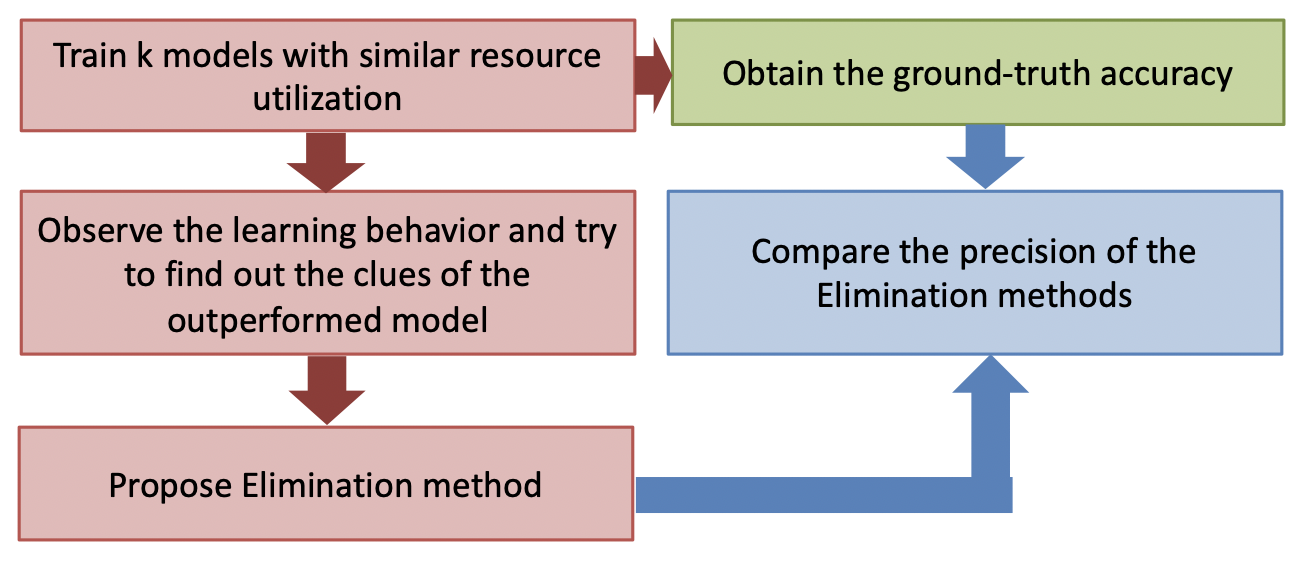}
\end{center}
   \caption{Steps to make up the hypothesis. \label{fig:hypothesis}}
\end{figure}

\begin{table}
	\vskip -0.1in     
\caption{Candidate models for observation.}     
  \vskip 0.05in
  \centering   

  \resizebox{1.0\columnwidth}{!}{ 
\begin{tabular}{|l|l|l|l|l|}
\hline
Candidate & \begin{tabular}[c]{@{}l@{}}Parameter \\ usage(M)\end{tabular} & Flops(M) & \begin{tabular}[c]{@{}l@{}}Top-1 Accuracy on\\  ImageNet ($350^{th}$ epoch)\end{tabular} & Ranking \\ \hline
C1        & 3.7                                                           & 296      & 75.65                                                                              & 4       \\ \hline
C2        & 3.8                                                           & 384      & 76.45                                                                              & 2       \\ \hline
C3        & 4.4                                                           & 314      & 76.14                                                                              & 3       \\ \hline
C4        & 4.7                                                           & 362      & 76.62                                                                              & 1       \\ \hline
\end{tabular}                                             
   }                                                                                 
	
\label{table:Cmw}
\end{table}

\begin{figure}[t]
\begin{center}

   \includegraphics[width=\linewidth]{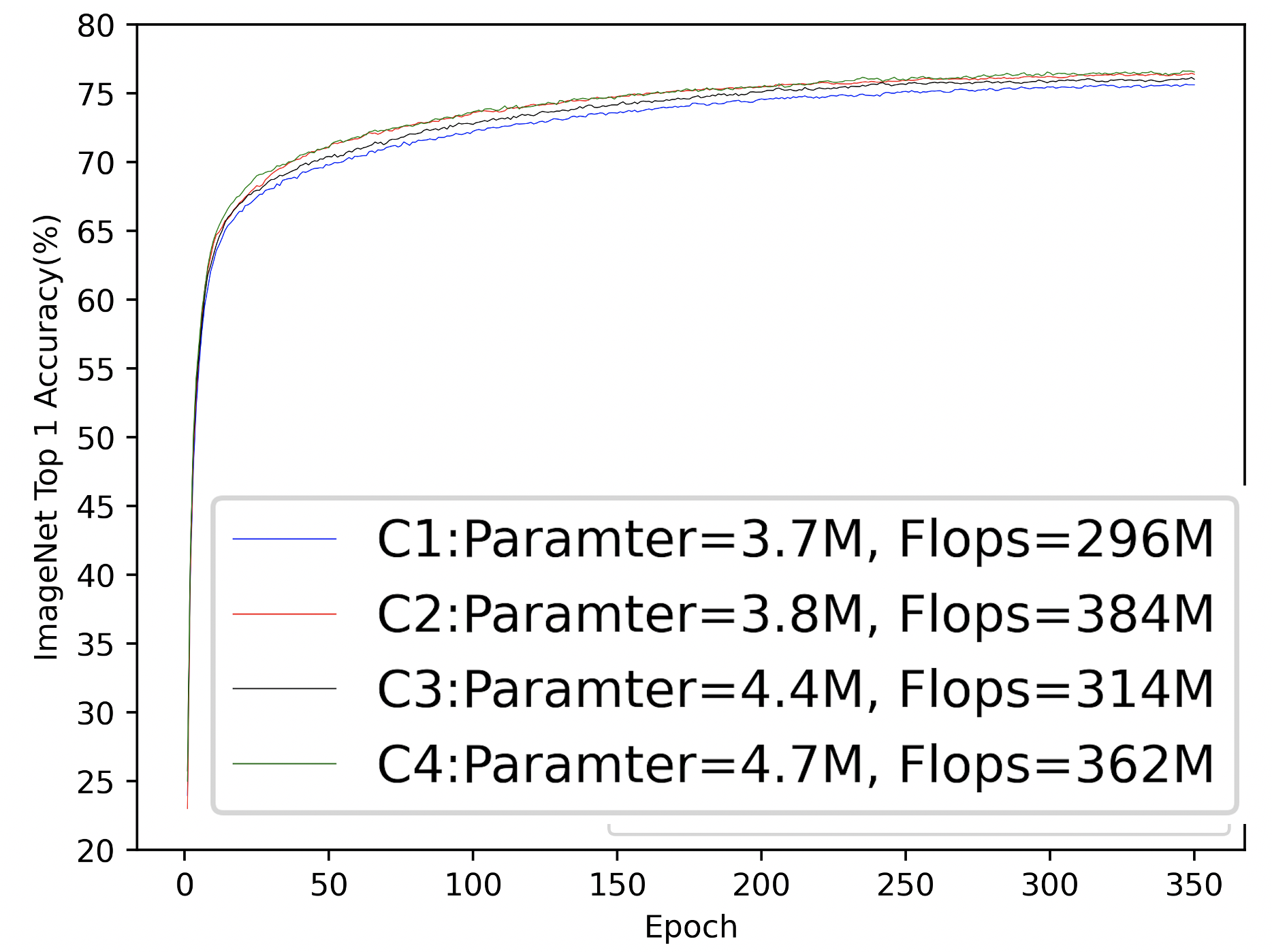}
\end{center}
   \caption{Accuracy curves for observation. \label{fig:acc_curve}}
\end{figure}

By the observation, we find that performance of the specific epoch is sometimes not representable of the final performance. As a result, we suggest that average accuracy is a better criteria for predicting outperformed model than using accuracy of specific epoch. There are two kinds of phenomenon of the accuracy curve. The first one is that the accuracy curve from a candidate is always higher than the other candidate's accuracy curve. Both methods of average and specific work perfectively to predict the outperformed CNN model. The other case is that the accuracy curve is crossing. The average method is possibly more accurate than the specific approach. Besides, the benefits of average performance is that the criteria is not arbitrary to the accuracy of specific epoch.

We evaluate the intuition by applying the criteria into the training results among the candidates. We divide 350 epochs as 35 rounds and 10 epochs per round as unit for observation. For each round, we observe the relative performance of candidate models, measuring how the extend of the criteria matches the finial performance. That is to say, whether the average accuracy or the specific accuracy can better reflect the final ranking. In equation \ref{eq:: specific_accuracy}, $Acc^{C_m}_{spe}(i)$ stands for specific accuracy of $i^{th}$ epoch on candidate $C_m$. $Rank_{spe}(r) = $ denotes an ascending order sequence  sorted by the accuracy of specific epoch (Round $r$) of candidate model.  $Mat_{spe}$ is for measuring how matching the criteria is for predicting the final ranking.

\begin{equation} \label{eq:: specific_accuracy} 
\begin{aligned}
\text{$Rank_{spe}(r)$} = & \  (C_1, C_2, C_3, C_4) \ if \  Acc^{C_1}_{spe}(10r) \ > ... > \ Acc^{C_4}_{spe}(10r) \\
\text{$Rank_{spe}(35)$} = & \  (C_1, C_3, C_2, C_4) \\
\text{$P_{spe}$} = &  \ \dfrac{ \displaystyle\sum_{r=1}^{35} Rank_{spe}(r)\ == \ Rank_{spe}(35)}{35} \\
\text{$P_{spe}$} = &  \ \dfrac{22}{35} \ \approx \ 63 \ \%
\end{aligned}
\end{equation}

In equation \ref{eq:: average_accuracy}, $Rank_{avg}(r)$ denotes an ascending order sequence but sorted by the average accuracy from first epoch to the current epoch $10r$. Using average accuracy  is 94\% matching the result.

\begin{equation} \label{eq:: average_accuracy} 
\begin{aligned}
\text{$Acc^{C_m}_{avg}(1,\ k)$} = &  \ \dfrac{ \displaystyle\sum_{i=1}^{k} Acc_{spe}^{C_m}(i)\ }{k} \\
\text{$Rank_{avg}(r)$} = & \  (C_1, C_2, C_3, C_4), \  \\
if \  Acc^{C_1}_{avg}(1, 10&r) > ... > Acc^{C_4}_{avg}(1, 10r) \\
\text{$P_{avg}$} = &  \ \dfrac{ \displaystyle\sum_{r=1}^{35} Rank_{avg}(r)\ == \ Rank_{spe}(35)}{35} \\
\text{$P_{avg}$} = &  \ \dfrac{33}{35} \ \approx \ 94 \ \%
\end{aligned}
\end{equation}

By the observation, the elimination criteria is based on sorted average accuracy, which is calculated form the first epoch to the current epoch. Comparing to the method from specific accuracy,  searching cost remains the same except for the average operation and maintaining the history of the accuracy. Therefore, we adopt the average accuracy as elimination criteria.

\subsection{Steps of algorithm of Network Candidate Search}

\begin{itemize}
	\setlength\itemsep{0em}
	\item STEP 1: Candidate initialization
		\subitem STEP 1.1: Defining candidate models
		\subitem STEP 1.2: Sorting candidates by resource usage
		\subitem STEP 1.3 : Dividing candidate into groups
	\item STEP 2: Training a round r of all survived candidates (We use e=10 epochs as a round)
	\item STEP 3: Eliminating models
		\subitem STEP 3.1: Calculating the average accuracy		
		\subitem STEP 3.2: Sorting average accuracy as ascending order
		\subitem STEP 3.3 : Eliminating half of the candidates with the sorted average accuracy per group
	\item STEP 4: Repeat 2 and 3 until finish the training and one candidate remains in the group respectively

\end{itemize}
\begin{table*}
	\vskip -0.1in     
	\caption{Comparison between state-of-the-art models and EfficientNet-eLite and EfficientNet-HF performance on ImageNet. The models with closing Top-1 accuracy on ImageNet are blocked together and organized as ascending order. Note that we define searching cost on ImageNet dataset counting from the start to eliminate until only one candidate survives per group.}     
  \vskip 0.05in
  \centering   

  \resizebox{0.9\textwidth}{!}{ 
\begin{tabular}{|c|c|c|c|c|c|} 
\hline

State-of-the-art models& Publication&Parameters(M)& Flops(M)&Top1 ACC.(\%)&Searching cost (GPU hours)\\ \hline

Mnas-small \cite{MnasNet}&CVPR 2019&1.9&65.1&64.9&40,000\\ 
MobileNet V3 small 0.75\cite{MobileNets_v3}&ICCV2019&2&{\bf44}&65.4&Manual\\ 
$Model(w_4, d_4,r_1)(eLite9)$&-&{\bf1.34}&74.56&{\bf65.46}&80\\ \hline \hline

GhostNet 0.5$\times$ \cite{GhostNet}&CVPR2020&2.6&{\bf42}&66.2&Manual\\ 
EfficientNet-HF5 &-&2.77&62.66&66.76&70\\ 
MobileNet V3 small 1.0\cite{MobileNets_v3}&ICCV2019&2.5&56&67.4&Manual\\ 
EfficientNet-HF4&-&2.98&65.55&67.95&40\\ \
$Model(w_4, d_2,r_1)(eLite8)$&-&{\bf1.6}&114.52&{\bf68.8}&100\\ \hline \hline

$Model(w_3, d_4,r_2)(eLite7)$&-&2.18&127.99&70.18&90\\ 
EfficientNet-HF3&-&3.44&92.29&70.32&40\\ \hline \hline

MobileNet\cite{MobileNets}&CVPR2017&4.24&569&70.6&Manual\\ 
CondenseNet (G=C=8)\cite{CondenseNet}&CVPR2018&2.9&274&71&Manual\\ 
MobileNet V2\cite{MobileNets_v2}&CVPR2018&3.4&300&72&Manual\\ 
$Model(w_3, d_3,r_1)(eLite6)$&-&{\bf2.52}&{\bf181.32}&{\bf72.23}&100\\ \hline \hline

ECA-Net\cite{ECA-Net}&CVPR2020&3.4&319&72.56&Manual\\ 
PeleeNet\cite{PeleeNet}&NeurIPS2018&2.8&508&72.6&Manual\\ 
DARTS\cite{DARTS}&ICLR2019&4.9&595&73.1&288\\ 
MobileNet V3 Large 0.75\cite{MobileNets_v3}&ICCV2019&4&{\bf155}&73.3&Manual\\ 
$Model(w_3, d_2,r_1)(eLite5)$&-&{\bf2.68}&206.72&{\bf73.33}&110\\ \hline \hline

EfficientNet-HF2 &-&2.77&246.86&73.51&100\\ \hline \hline

ShuffleNet 2$\times$ (g=3) \cite{ShuffleNet}&CVPR2017&5.2&524&73.7&Manual\\ 
GhostNet 1.0$\times$ \cite{GhostNet}&CVPR2020&5.2&{\bf141}&73.9&Manual\\ 
NASNet-A\cite{NASNet}&CVPR2018&5.3&508&74&48,000\\ 
AmoebaNet-B (N = 3, F = 62) \cite{AmoebaNet}&AAAI2019&5.3&555&74&75,600\\ 
PNASNet \cite{PNAS}&ECCV2018&5.1&588&74.2&3,600\\ 
GreedyNAS (latency$\leq$ 80ms) \cite{GreedyNAS}&CVPR2020&4.1&324&74.39&-\\ 
$Model(w_2, d_3,r_2)(eLite4)$&-&{\bf3.17}&231.88&{\bf74.41}&110\\ \hline \hline

AmoebaNet-A (N = 4, F = 50) \cite{AmoebaNet}&AAAI2019&5.1&555&74.5&75,600\\ 
Proxyless \cite{Proxyless}&ICLR2019&4&320&74.6&200\\ 
GreedyNAS (FLOPs$\leq$ 322M) \cite{GreedyNAS}&CVPR2020&3.8&320&74.85&-\\ 
FBNet-C \cite{FBNet}&CVPR2019&5.5&375&74.9&216\\ 
EfficientNet-HF1 &-&{\bf3.17}&{\bf296.37}&{\bf75.08}&100\\ \hline \hline

MnasNet-A1\cite{MnasNet}&CVPR2019&3.9&315&75.2&40,000\\ 
MobileNet V3 Large 1.0\cite{MobileNets_v3}&ICCV2019&5.4&{\bf219}&74.9&Manual\\ 
$Model(w_2, d_2,r_1)(eLite3)$&-&{\bf3.78}&296.88&{\bf75.65}&130\\ \hline \hline

AmoebaNet-C (N = 4, F = 50) \cite{AmoebaNet}&AAAI2019&6.4&570&75.7&75,600\\ 
GhostNet 1.3$\times$ \cite{GhostNet}&CVPR2020&7.3&{\bf226}&75.7&Manual\\ 
$Model(w_1, d_3,r_1)(eLite2)$&-&{\bf4.42}&313.86&{\bf76.14}&150\\ \hline \hline

EfficientNet-HF0&-&3.81&383.49&76.46&100\\ 
$Model(w_1, d_2,r_1)(eLite1)$&-&4.74&362.62&76.62&-\\ 
$Model(w_1, d_1,r_1)(eLite0)$&-&5.33&385.81&76.89&-\\ 
EfficientNet-B0 \cite{EfficientNet}&ICML2019&5.3&390&77.3&-\\ \hline 
\end{tabular}                                                     
}                                                                                 

\label{table:state}  
\end{table*} 

\section{Hardware friendly EfficientNet}
\label{c:proposed}
In this section, applying the same scaling coefficients in Table~\ref{table:result_of_scaling}, we consider hardware friendly designs into the scaled-down models. Meanwhile, the hardware-friendly candidate pool can be determined, using NCS, which features for generalizing for any kind or any type of neural network, to select the outperformed models. The hardware-friendly designs can be categorized by channels and input resolution.

\subsection{Adjustment for input resolution}
\label{section:Proposed friendly design}

Two kinds of input resolution are provided, $256\times256$ and $128\times128$. The size of feature map for each operator becomes the power of two number, which is benefits for the data partitioning problem. Besides, the operation of division (average pooling) can be done by shifting rather than the hardware division unit.
\subsection{Compound channels rounding}
\label{section:Compound channels rounding}
The basic idea is to make each channels to be power of two as well as to avoid distorting the model structure as much as possible. Denote the original channels $C_s$ on the stage s. Channel by rounding up $C^{RU}_s$ is defined as $2^{ceiling(log_2C_s)}$. Channel by rounding down $C^{RD}_s$ is defined as $2^{floor(log_2C_s)}$. Compound rounding $C_s^{CR}$, defined in Equation \ref{eq:CR}, is combined rounding up and down for less adjustment of channels because we want to keep the original shape of EfficientNet~\cite{EfficientNet}.

\begin{equation} \label{eq:CR} 
\begin{aligned}
C_{s}^{CR} = \left\{
             \begin{array}{lr}
             C^{RU}_s &   \text{, if } C^{RU}_s - C_s  <  C_s  - C^{RD}_s \\
             C^{RD}_s &  \text{, else} \\
             \end{array}
	    \right.
\end{aligned}
\end{equation}
\begin{figure}[t]
\begin{center}

   \includegraphics[width=\linewidth]{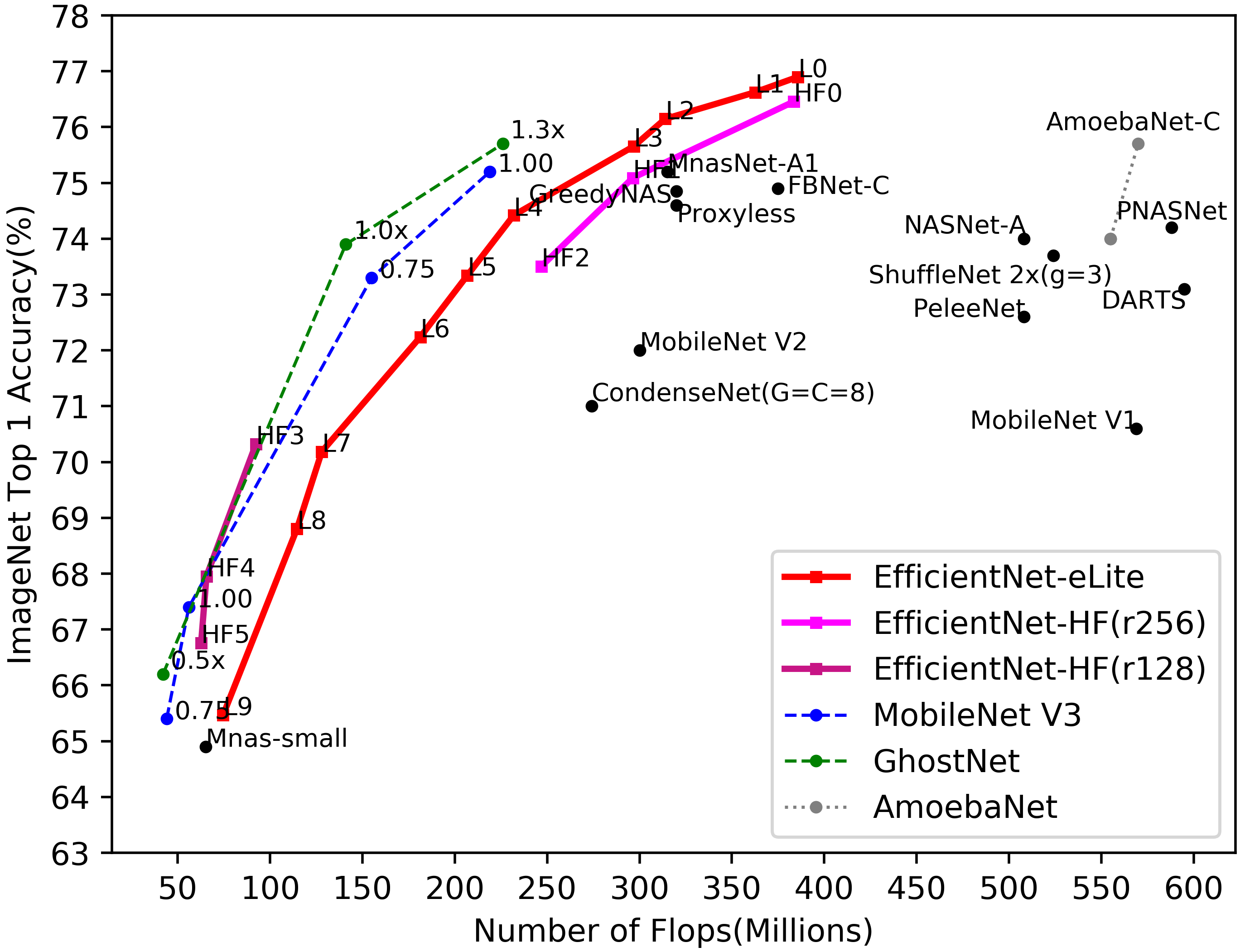}
\end{center}
   \caption{The performance of Flops and Top-1 accuracy on ImageNet.}
\label{fig:Flops}
\end{figure}

\section{Experimental Results}
\label{c:proposed}

\subsection{Implementation environment}
\label{section:Implementation details}

Our experiment is conducted using NVIDIA 2080Ti and i7-9700K CPU with Pytorch implementation. The training parameter settings are as follow, which is fundamentally following EfficientNet\cite{EfficientNet} except for some GPU equipment limitation. We adopt the batch size to be 100 and the training epoch to be 350. The Optimization is based on RMSProp. The Data augmentation policy is adopted from \cite{Fast_Autoaugmentation}.

\subsection{Results from each group}
\label{section:Result from each group}

The champion model of each group is represented as a member of EfficientNet-eLite and EfficientNet-HF and we give the lager id for the more lightweight member. We find that the winner of each group has the resolution coefficient either $r_1$ or $r_2$. Because we group CNN models by similar parameter usage and Flops, we realize that models with higher input information tend to outperform others under the similar resource usage condition when models are scaled down.

\subsection{State-of-the-art models on ImageNet}
\label{section:State-of-the-art model on ImageNet}

Figure \ref{fig:Parameter} shows the model size with the ImageNet Top-1 accuracy. Under the same accuracy, our proposed models are smaller than the other state of the arts. From the vertical perspective for same parameter usage, our proposed models have higher Top-1 accuracy on ImageNet than the other state of the arts.
Figure \ref{fig:Flops} shows the Flops with the ImageNet Top-1 accuracy. Our proposed models outperform the most of the state of the arts with fewer floating point operation and higher accuracy.

\section{Conclusion}
\label{c:conclusion}

A family of extremely lightweight CNN models for edge devices is proposed. Particularly, EfficientNet-eLite 9 is more lightweight than the best and smallest existing model Mnas-small\cite{MnasNet}. We study the trade-off between hardware resource and accuracy by a novel of Network Candidate Search, which candidates are determined by scaling down EfficientNet. We find that scaling down width and depth tends to have less accuracy drop than reducing the input information. Besides, grouping and elimination concept are introduced for effectively selecting the potential structure and reducing searching cost at the same time. We propose using average accuracy to speculate the potential CNN models. To push the state-of-the-art CNN embedded for edge application, we also propose hardware-friendly CNN models by using the same methodology of NCS along with hardware-friendly adjustments. Finally, both of the families of models outperform the state-of-the-art CNN models with less parameter usage and higher accuracy on ImageNet. 
\newpage
{\small
\bibliographystyle{ieee_fullname}
\bibliography{egbib}
}

\end{document}